\documentclass[conference]{IEEEtran}

\makeatletter

\def\ps@IEEEtitlepagestyle{%
  \def\@evenfoot{}%
}

\usepackage{blindtext}
\usepackage{eso-pic}
\IEEEoverridecommandlockouts
\usepackage{cite}
\usepackage{amsmath,amssymb,amsfonts}
\usepackage{algorithmic}
\usepackage{graphicx}
\usepackage{textcomp}
\usepackage{xcolor}
\def\BibTeX{{\rm B\kern-.05em{\sc i\kern-.025em b}\kern-.08em
    T\kern-.1667em\lower.7ex\hbox{E}\kern-.125emX}}
    
\usepackage{eso-pic}
\newcommand\AtPageUpperMyright[1]{\AtPageUpperLeft{%
 \put(\LenToUnit{0.17\paperwidth},\LenToUnit{-2cm}){%
     \parbox{0.9\textwidth}{\raggedleft\fontsize{8}{11}\selectfont #1}}%
 }}%
\newcommand{\conf}[1]{%
\AddToShipoutPictureBG*{%
\AtPageUpperMyright{#1}
}
}

\begin{document}
\title{\vspace*{1cm} A TinyML Reinforcement Learning Approach for Energy-Efficient Light Control in Low-Cost Greenhouse Systems\\

\thanks{This research is based upon work supported by North Dakota State University and the U. S. Department of Agriculture, Agricultural Research Service, under agreement No. 58-6064-3-011.}
}

\author{\IEEEauthorblockN{Mohamed Abdallah Salem}
\IEEEauthorblockA{
\textit{North Dakota State University}\\
Fargo, USA \\
Mohamed.Salem@ndsu.edu}
\and
\IEEEauthorblockN{Manuel Cuevas Perez}
\IEEEauthorblockA{
\textit{North Dakota State University}\\
Fargo, USA \\
Manuel.Cuevasperez@ndsu.edu}
\and
\IEEEauthorblockN{Ahmed Harb Rabia}
\IEEEauthorblockA{
\textit{dept. Agricultural and Biosystems Engineering} \\
\textit{North Dakota State University}\\
Fargo, USA \\
Ahmed.Rabia@ndsu.edu}
}

\maketitle
\conf{\textit{ Accepted for publication in the Proceedings of the 5. Interdisciplinary Conference on Electrics and Computer (INTCEC 2025) \\ 
15-16 September 2025, Chicago-USA}}

\begin{abstract}
This study presents a reinforcement learning (RL)-based control strategy for adaptive lighting regulation in controlled environments using a low-power microcontroller. A model-free Q-learning algorithm was implemented to dynamically adjust the brightness of a Light-Emitting Diode (LED) based on real-time feedback from a light-dependent resistor (LDR) sensor. The system was trained to stabilize at 13 distinct light intensity levels (L1 to L13), with each target corresponding to a specific range within the 64-state space derived from LDR readings. A total of 130 trials were conducted, covering all target levels with 10 episodes each. Performance was evaluated in terms of convergence speed, steps taken, and time required to reach target states. Box plots and histograms were generated to analyze the distribution of training time and learning efficiency across targets. Experimental validation demonstrated that the agent could effectively learn to stabilize at varying light levels with minimal overshooting and smooth convergence, even in the presence of environmental perturbations. This work highlights the feasibility of lightweight, on-device RL for energy-efficient lighting control and sets the groundwork for multi-modal environmental control applications in resource-constrained agricultural systems.
\end{abstract}

\begin{IEEEkeywords}
Autonomous control, greenhouse automation, light intensity regulation, precision agriculture, Q-learning, reinforcement learning, temperature control
\end{IEEEkeywords}

\section{Introduction}
Greenhouses have become a cornerstone of modern agriculture, providing controlled environments that mitigate the uncertainties of outdoor conditions while enabling higher crop yields and improved year-round productivity. The ability to precisely regulate environmental parameters within these structures is not only crucial for maximizing agricultural output but also for addressing the broader societal challenge of using energy resources efficiently \cite{paris2022energy}. 

Maintaining optimal conditions for plant growth within Controlled Environment Agriculture (CEA) systems stands as a fundamental objective in this endeavor. The increasing demand for food due to population growth, coupled with the growing instability of traditional farming methods caused by climate variability, underscores the critical need for advanced greenhouse control systems capable of enhancing both productivity and resource management \cite{engler2021review}. 

Furthermore, the energy-intensive nature of greenhouse operations makes the pursuit of efficient energy consumption not just an economic necessity for growers but also a significant contribution to global sustainability efforts \cite{saadi2025comparative}. Among the many environmental parameters regulated in a greenhouse, lighting is one of the most influential, as it affects photosynthesis, energy usage, and overall plant development \cite{liu2021light}. Precise management of lighting, particularly in settings relying on supplemental artificial sources, is crucial for optimizing energy costs and ensuring consistent plant quality \cite{jayalath2024energy}.

Conventional lighting control mechanisms---such as fixed schedules or simple rule-based strategies---often fail to adapt to the dynamic conditions of greenhouse environments, such as shifting cloud cover or plant growth stages \cite{afzali2022optimal}. Reinforcement learning (RL), which enables agents to learn optimal control policies through direct interaction with their environment, offers a promising alternative \cite{mallick2024reinforcement}. The urgency to improve food security amid global population growth and increasing climate variability is driving interest in Controlled Environment Agriculture (CEA), where precise environmental regulation (light, temperature, humidity, and CO$_2$ levels) is crucial for optimizing crop productivity and resource use \cite{ullah2024global, benitez2023enhancing}.

Traditional greenhouse climate control frequently relies on Proportional-Integral-Derivative (PID) controllers due to their simplicity and widespread use. However, PID systems require manual tuning and often fail under non-linear, time-varying conditions caused by environmental disturbances or plant developmental stages \cite{mengesha2025comparison}. Adaptive controllers, such as the one proposed by \cite{kim2024model,morcego2023reinforcement}, use radial basis function (RBF) neural networks to dynamically tune PID gains in response to changing greenhouse conditions and have shown improved adaptability under disturbance-laden environments \cite{zeng2012adaptive}. Despite their effectiveness, these models are computationally demanding and not well suited for low-power embedded systems.

Recent studies have begun to integrate RL frameworks with classical and model-based controllers to improve robustness and adaptability. For example, RL-guided Model Predictive Control (MPC) has outperformed traditional control strategies in both deterministic and uncertain greenhouse scenarios \cite{msaad2025rlmpc}. Additionally, \cite{platero2024enhancing} demonstrated the effectiveness of RL-enhanced Internet of Things (IoT) systems for autonomous greenhouse climate regulation, particularly in energy savings and reduced human intervention \cite{platero2024iot}.

However, most of these RL-driven solutions rely on high-performance edge devices or cloud-based processing, making them impractical for deployment in cost-sensitive or remote agricultural settings. To address this, the emerging field of Tiny Reinforcement Learning (TinyRL) adapts lightweight RL algorithms, such as tabular Q-learning, for use on microcontrollers like the ESP32 \cite{ren2021tinyol, szydlo2022tinyrl}. These implementations enable on-device decision-making with limited memory and computational resources.

In this study, we propose a lightweight embedded RL solution for adaptive lighting control in greenhouses. The system is implemented using an ESP32 microcontroller, a light sensor (LDR), and a controllable LED module. Unlike computationally expensive deep RL methods, our solution employs both tabular and fuzzy Q-learning to learn optimal lighting strategies on-device, framed as a Markov Decision Process (MDP). This framework enables dynamic response to environmental changes while avoiding the energy and communication overhead of cloud dependency. The paper details the system architecture, training process, evaluation methodology, and comparative analysis, demonstrating how lightweight RL can be practically deployed for precise, efficient greenhouse lighting control.

\section{Materials and Methods}

This section provides a comprehensive overview of the design, implementation, and testing of the proposed reinforcement learning (RL) system for greenhouse lighting control. It includes the conceptual proposal, system architecture, hardware integration, and the step-by-step development of the RL model. The methodology is divided into two main parts: general RL implementation principles and the lighting control system.

\subsection{General Proposal and RL Framework}

The primary goal of this project is to implement a fully autonomous RL-based system capable of maintaining optimal lighting levels in a greenhouse environment. This involves designing an agent for light, integrating real-time data collection, and enabling the system to respond dynamically to environmental fluctuations. We used a tabular Q-learning algorithm for the agent due to its simplicity and suitability for embedded platforms. The algorithm maps discrete environmental states to specific actions, learning an optimal policy over time through reward feedback. Each agent maintains its own Q-table, which is updated based on the state-action-reward-next state tuple.

The RL system was designed and implemented at an experimental scale using ESP32 Development board micro-controller. Data was collected in real-time from the sensors, logged through serial output, and used for policy learning and evaluation. 
To simulate dynamic environments, artificial perturbations were introduced during training (e.g., turning off external lights). These actions exposed the agent to real-world conditions, helping it learn to adaptive behaviors rather than static responses. After training, the agents Q-table reflected a preference for actions that maximized long-term rewards.

\begin{figure}[htbp]
\centering 
\includegraphics[width=\columnwidth]{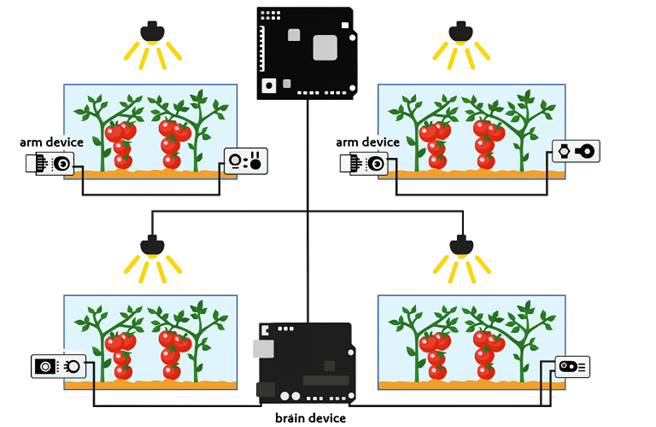}
\caption{Distributed Reinforcement Learning-Based Lighting Control in Greenhouse Units.}
\label{setup}
\end{figure}

The diagram on Fig.~\ref{setup} illustrates a modular greenhouse setup where each unit is equipped with an edge device consisting of a light-dependent resistor (LDR) sensor and light emitting diode (LED) lighting system. These devices operate under the control of a central brain device that coordinates distributed Q-learning policies for adaptive lighting. The system continuously monitors ambient light levels and adjusts LED brightness in real time to maintain optimal illumination conditions for plant growth.

\begin{figure}[ht]
\centering
\includegraphics[width=\columnwidth]{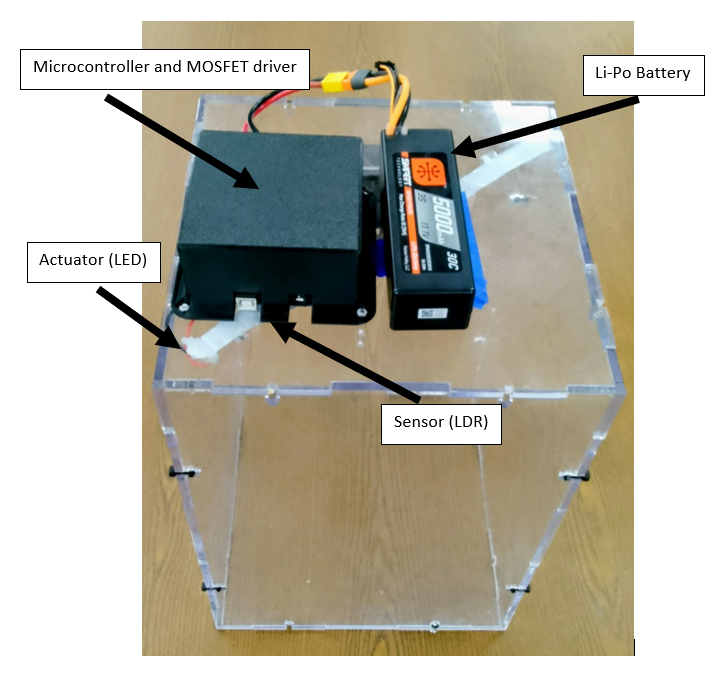}
\caption{Experimental test setup showing the main components: LED (actuator), LDR (sensor), microcontroller (controller), and power source.}
\label{fig:setup}
\end{figure}

The physical test setup used to conduct the experiments is shown in Figure~\ref{fig:setup}. It includes an actuator (LED), sensor (LDR), microcontroller with MOSFET driver, and a Li-Po battery, all mounted within a controlled transparent chamber.

\subsection{RL Model Construction}

The reinforcement learning agents developed for this study are based on the classical Q-learning algorithm, a value-based, model-free method for solving Markov Decision Processes (MDPs) \cite{watkins1992q, sutton2018reinforcement}. The objective of the agent is to learn an optimal policy that maps environmental states to actions in a way that maximizes the expected cumulative long-term reward. Below are the core mathematical components of the Q-learning framework that guide this learning process.

\subsubsection{State and Action Representation}

Let the environment be defined by a finite set of states $S = \{s_1, s_2, \ldots, s_n\}$ and a finite set of actions $A = \{a_1, a_2, \ldots, a_n\}$. At each time step $t$, the agent observes the current state $s_t$, selects an action $a_t$, transitions to a new state $s_{t+1}$, and receives a reward $r_t$.

\subsubsection{Q-Value Update Rule}
The core learning mechanism is governed by the Bellman equation (Sutton and Barto n.d.) used to iteratively update the Q-values stored in a table $Q(s,a)$:

\begin{multline}
    Q(s_t, a_t) \leftarrow Q(s_t, a_t) + \\
    \alpha \left[ r_t + \gamma \max_{a'} Q(s_{t+1}, a') - Q(s_t, a_t) \right]
\end{multline}

where $\alpha = \text{learning rate } (0 < \alpha \leq 1)$, which controls how much newly acquired information overrides old information. \\
$\gamma = \text{discount factor } (0 \leq \gamma \leq 1)$, determining the importance of future rewards. \\
$r_t = \text{the reward received after taking action } a_t \text{ in state } s_t$. \\
$\max_{a'} Q(s_{t+1}, a') = \text{best future value from the next state.}$

\subsubsection{Action Selection Strategy}

During training, the agent follows an $\varepsilon$-greedy policy to balance exploration and exploitation:
$$
a_t = \begin{cases}
    \text{random action} & \text{with probability } \varepsilon \\
    \arg \max Q(s,a) & \text{with probability } (1-\varepsilon)
\end{cases}
$$

The exploration rate $\varepsilon$ (epsilon) starts high (e.g., 0.5) and decays over time, enabling the agent to initially explore different actions and gradually converge to exploiting the best-known policy.

\subsubsection{Reward Selection Strategy}

The reward function is a sparse binary function used to reinforce desirable states. For example, for the lighting system:
$$
r_t = \begin{cases}
    +1 & \text{if } s_t = s_{\text{target}} \\
    -1 & \text{otherwise}
\end{cases}
$$

For the proposed approach, the following hyperparameters were empirically determined and utilized:

\begin{table}[htbp] 
\caption{Hyperparameters used in the Proposed Approach}
\label{tab:hyperparameters}
\begin{center} 
\small 
\begin{tabular}{|l|c|c|} 
\hline
\textbf{Parameter Name} & \textbf{Symbol} & \textbf{Value} \\ 
\hline
Learning Rate      & $\alpha$        & $0.1$          \\
\hline
Discount Factor    & $\gamma$        & $0.9$          \\
\hline
Exploration Rate   & $\varepsilon$   & $0.5$          \\
\hline
\end{tabular}
\end{center}
\end{table}

\section{Results and Discussion}

The reinforcement learning agent was trained across 13 different target lighting levels (denoted as L1 through L13), each corresponding to a discrete state range within the 64-state representation derived from LDR sensor input. For each target level, 10 independent training trials were conducted, resulting in a total of 130 experimental runs. The results were analyzed in terms of convergence behavior, training time, and number of steps required to stabilize at the target lighting level.

\begin{figure}[ht]
    \centering
    \includegraphics[width=\columnwidth]{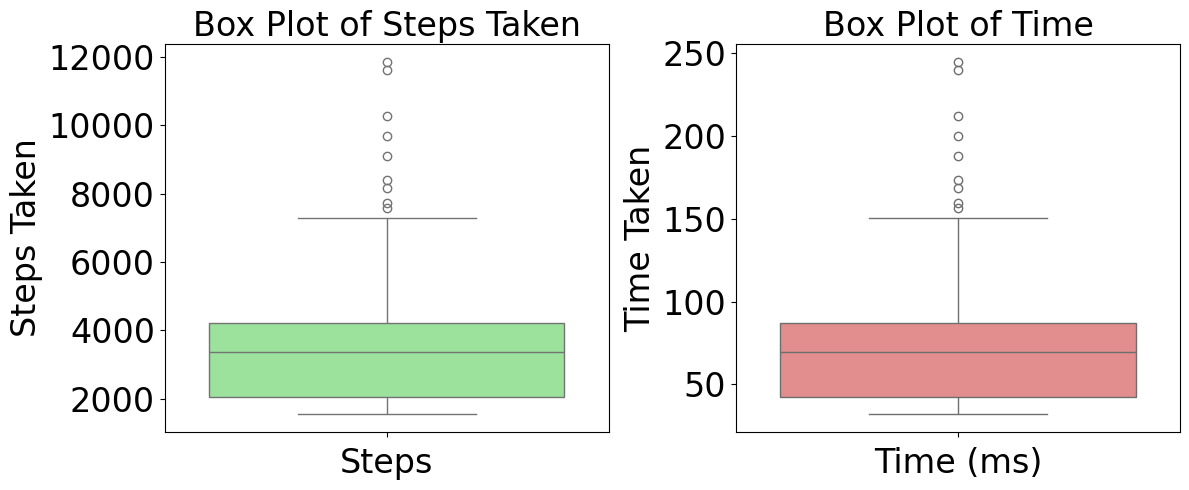}
    \caption{Box plots summarizing the distribution of steps taken (left) and time required (right) to reach each target lighting level across 130 training episodes.}
    \label{fig:BoxPlots}
\end{figure}

Figure~\ref{fig:BoxPlots} presents box plots summarizing the number of steps taken and the time (in milliseconds) required to converge to the target state during the training process. The left plot shows the distribution of steps taken across all 130 runs, revealing a median around 3300 steps, with some outliers requiring significantly more steps. The right plot illustrates a similar distribution for convergence time, with a median below 75 milliseconds. Both plots demonstrate that while most training runs converge efficiently, a subset of cases exhibit extended learning durations, likely due to challenging state transitions or local minima.

\begin{figure}[ht]
    \centering
    \includegraphics[width=\columnwidth]{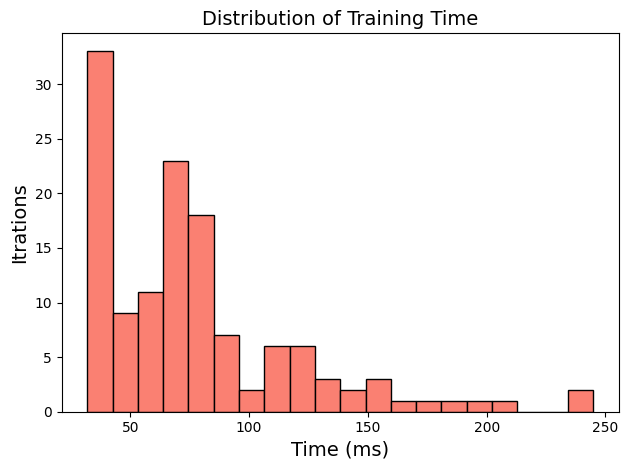}
    \caption{Histogram showing the distribution of training time over all episodes. The majority of runs converged in under 100 ms.}
    \label{fig:Histogram}
\end{figure}

The histogram in Figure~\ref{fig:Histogram} further illustrates the distribution of training time. Most episodes converged in under 100 ms, indicating the suitability of this approach for real-time embedded lighting control. The peak around 35--50 ms highlights the efficiency of the learning model in simpler target ranges, while a long tail reflects more complex or delayed episodes.

\begin{figure}[ht]
    \centering
    \includegraphics[width=\columnwidth]{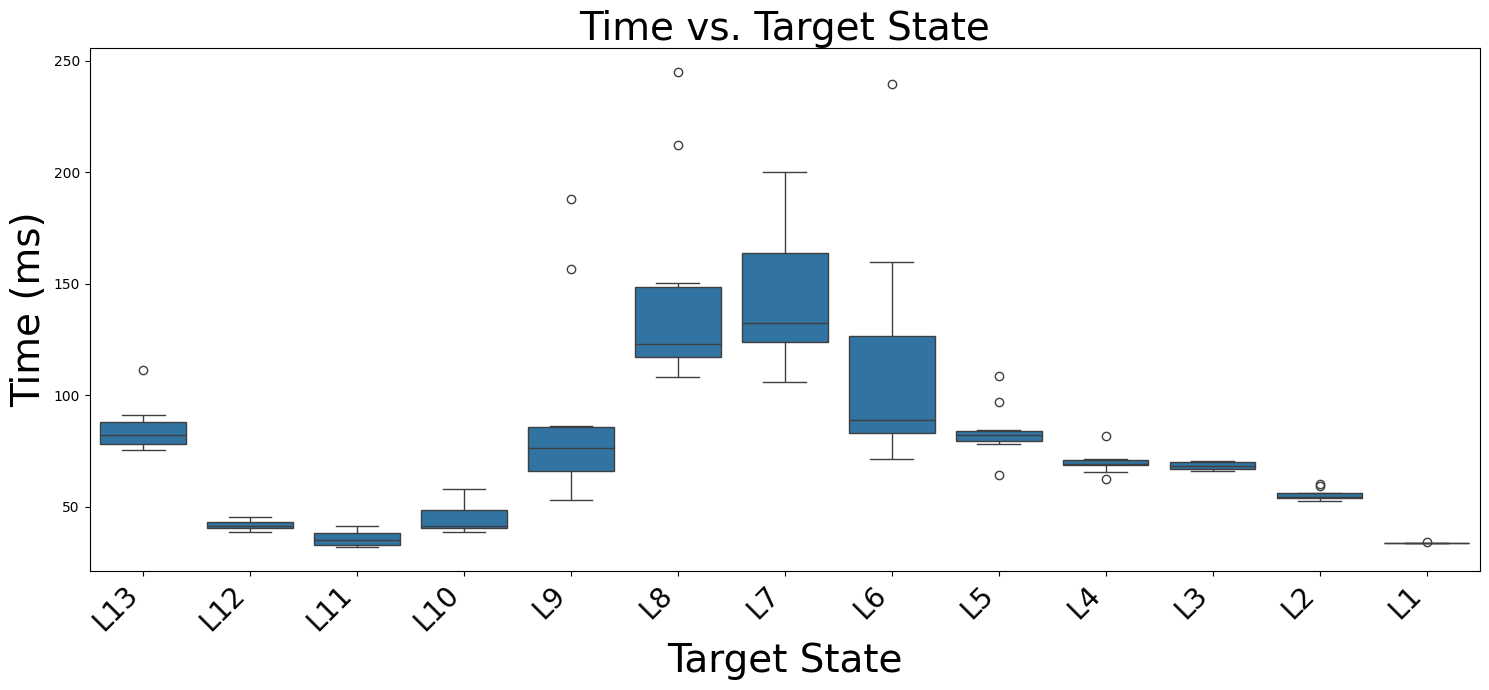}
    \caption{Convergence time to each target lighting level (L1 to L13). Intermediate levels generally required more time due to ambiguous boundaries.}
    \label{fig:BoxPerLevel}
\end{figure}

To understand the agent's performance with respect to individual target levels, Figure~\ref{fig:BoxPerLevel} shows box plots of convergence time grouped by target state level (L1 to L13). Results reveal that middle-range targets (e.g., L6 to L8) took longer on average to converge compared to extreme levels (L1 or L13), possibly due to their placement between multiple decision thresholds or due to less distinguishable sensor dynamics.

\begin{figure}[ht]
    \centering
    \includegraphics[width=\columnwidth]{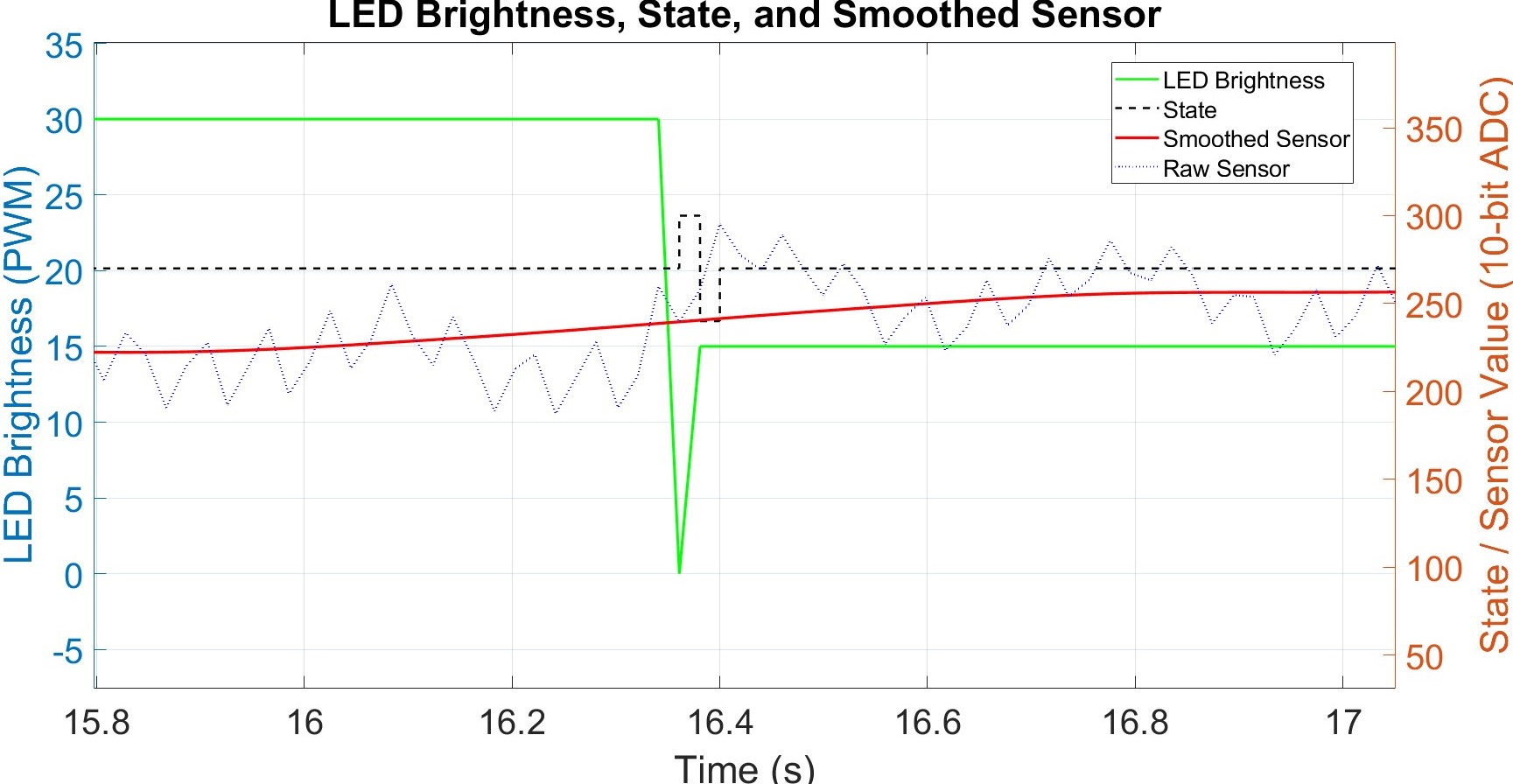}
    \caption{Zoomed sensor response, LED brightness, and state transitions during a single target trial.}
    \label{fig:ZoomPlot1}
\end{figure}

\begin{figure}[ht]
    \centering
    \includegraphics[width=\columnwidth]{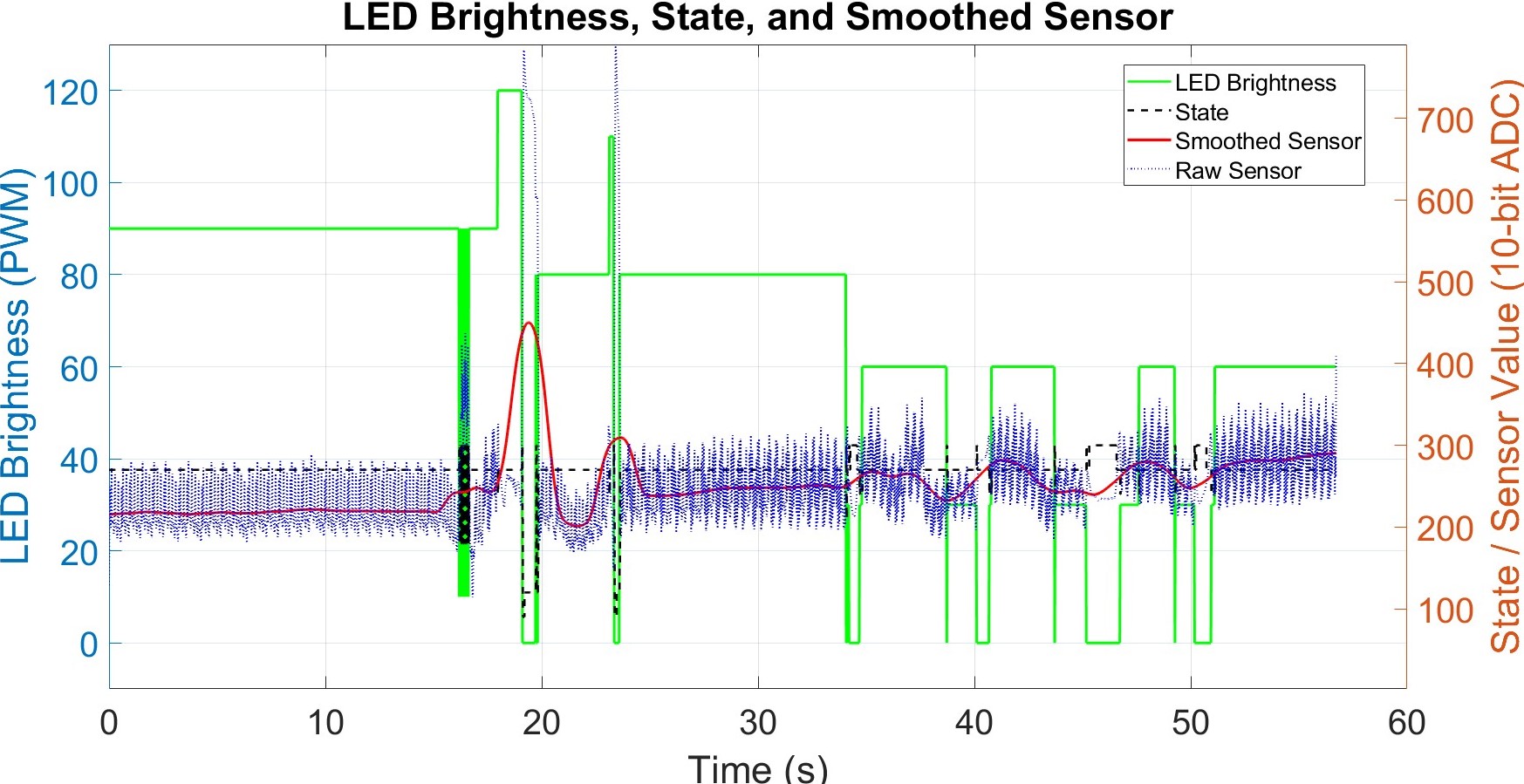}
    \caption{System performance in a variable lighting scenario with real-time adaptation.}
    \label{fig:ZoomPlot2}
\end{figure}

\begin{figure}[ht]
    \centering
    \includegraphics[width=\columnwidth]{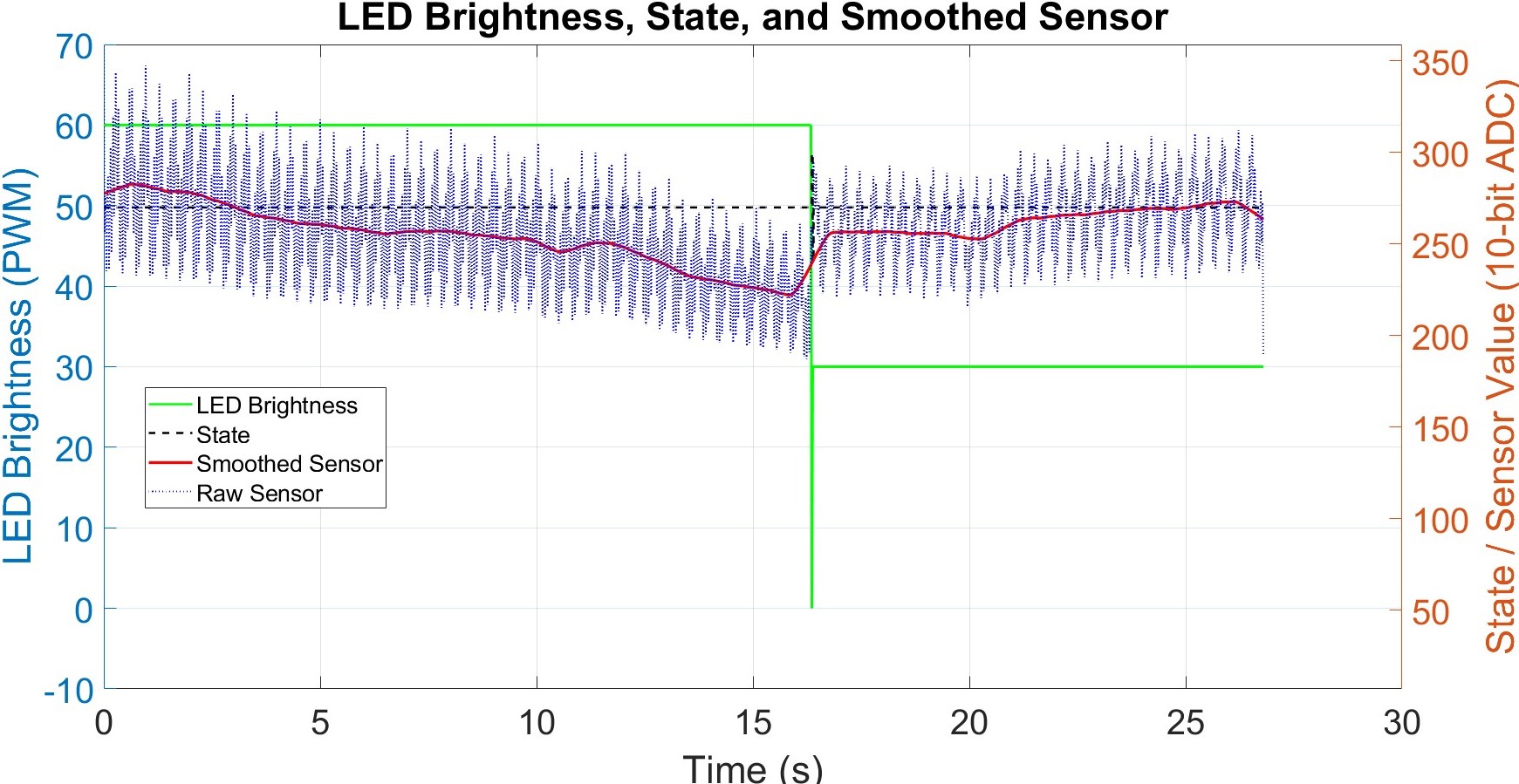}
    \caption{System response to a step change in environment, demonstrating rapid and stable adjustment.}
    \label{fig:ZoomPlot3}
\end{figure}

Figures~\ref{fig:ZoomPlot1} through~\ref{fig:ZoomPlot3} provide detailed examples of actual sensor response behavior during convergence. The plots overlay raw LDR sensor readings, smoothed curves, LED brightness (PWM signal), and discrete agent states. These visualizations validate that the system is able to adjust LED output and stabilize sensor readings to fall within the desired target range over time. The smoothed curve closely tracks the raw signal while filtering out high-frequency noise, indicating robust environmental response.

Figure~\ref{fig:ZoomPlot2} illustrates the dynamic behavior of the RL-based lighting controller under varying environmental conditions. The figure plots LED brightness (green), discrete RL state (black dashed), smoothed sensor readings (red), and raw LDR readings (blue dotted) over a 60-second interval. Around the 17--20 second mark, a pronounced spike in the sensor reading is observed, rising from a baseline near 260 ADC units to over 650 units within 2--3 seconds. This abrupt increase, followed by a sharp decay, reflects a strong external disturbance---such as an external light source or a reflective object---temporarily affecting the sensor's exposure.

Despite this disturbance, the RL agent responds by quickly adjusting the LED brightness to the minimum (PWM = 0), attempting to counteract the sudden rise in illumination. The system then re-evaluates the state and applies a new action, seen in the step-wise recovery between 20--25 seconds, as the smoothed sensor reading settles toward the target zone. The smoothed signal plays a critical role in ensuring that short-term fluctuations do not result in overly aggressive actions, thereby preventing oscillations.

Between 35--55 seconds, multiple smaller disturbances cause several fluctuations in the raw sensor readings. These correspond to physical changes in the environment such as hand movement, object shadowing, or flickering background light. Each of these events is reflected in the state's response and the corresponding LED brightness adjustments. Notably, despite at least five sudden dips and surges in the raw sensor data, the smoothed reading remains more stable-oscillating within a narrow band between 280--310 units. The RL system reacts with PWM adjustments ranging from 0 to 80, showcasing robustness to noise while continuing to regulate the target state.

\begin{figure}[ht]
    \centering
    \includegraphics[width=\columnwidth]{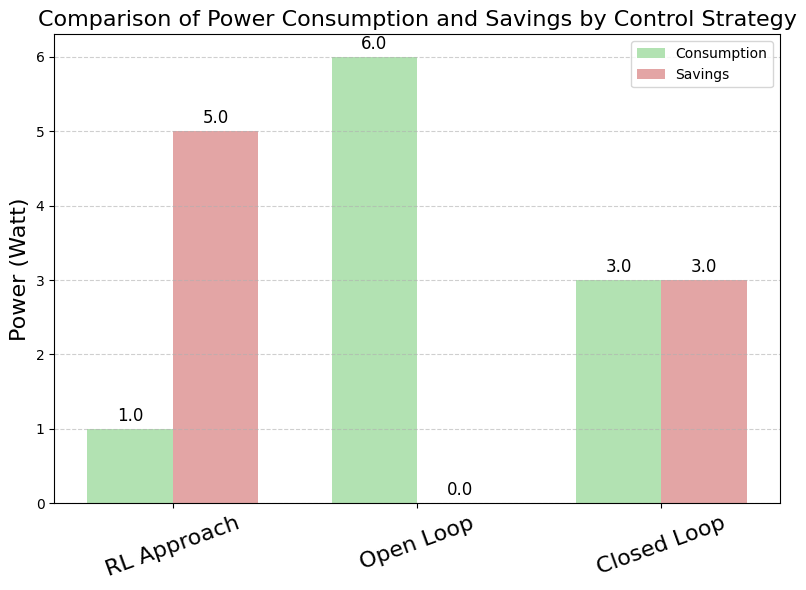}
    \caption{Comparison of power consumption and energy savings among RL-based, open-loop, and closed-loop lighting control strategies.}
    \label{fig:power_comparison}
\end{figure}

Figure~\ref{fig:power_comparison} compares the energy consumption and savings achieved using three different lighting control strategies: the reinforcement learning (RL) approach, an open-loop system, and a closed-loop feedback system. The RL approach demonstrated superior energy efficiency, consuming only 1 watt while achieving a significant energy saving of 5 watts. In contrast, the open-loop system, which lacks any form of feedback or adaptive control, resulted in the highest energy consumption at 6 watts and failed to achieve any energy savings. The closed-loop system offered a more balanced performance, with both consumption and savings at 3 watts, indicating its ability to respond to real-time sensor input, albeit without optimization. These findings clearly highlight the benefits of employing learning-based adaptive control, where the RL agent learns to minimize energy usage while maintaining the desired lighting conditions, outperforming both conventional fixed-output and reactive systems.

Although tabular Q-learning offers a simple and interpretable framework for reinforcement learning in discrete environments, it has several limitations. One major drawback is its inability to scale efficiently to high-dimensional state or action spaces. As the number of states grows, the memory required to store the Q-table increases rapidly, leading to inefficiencies in both storage and learning speed. Additionally, tabular methods lack generalization - each state-action pair must be learned independently, with no sharing of knowledge between similar states. This makes learning slow and data-inefficient, particularly in dynamic or partially observable environments. In practical applications involving continuous control or complex sensor readings, function approximation techniques such as Deep Q-Networks (DQNs) are often preferred, as they can handle large or continuous spaces and enable better generalization across unseen states, but such techniques require higher computational resources and larger memory.

While the current Q-learning implementation demonstrates effective light regulation within a discrete 64-state environment, scalability remains a critical limitation of tabular methods. As the number of states and actions increases, the size of the Q-table grows exponentially, resulting in significant memory usage and slower convergence. For example, extending the current setup to incorporate additional environmental variables (e.g., temperature, humidity, motion) would drastically expand the state-action space, making it impractical to store and update Q-values in a flat table format. Moreover, tabular methods do not generalize across unseen states, requiring each state-action pair to be explicitly visited and updated through experience. These factors limit the applicability of traditional Q-learning to more complex or continuous control problems.

Overall, the results demonstrate the viability of reinforcement learning for adaptive lighting control in real-time embedded systems. The system efficiently learns optimal policies with limited memory, and can adapt to various environmental conditions with consistent convergence behavior. Future work may extend this model to support temperature and humidity control using the same RL framework.

\section{Conclusion}

This work successfully demonstrated a reinforcement learning (RL)-based approach for real-time adaptive lighting control using low-cost embedded hardware. By discretizing the environment into 64 states and training the agent across 13 distinct target illumination levels, the system learned to autonomously adjust LED brightness based on feedback from a light-dependent resistor (LDR). The Q-learning algorithm, implemented onboard the microcontroller, proved capable of optimizing control actions in response to dynamically changing environmental light levels.

Extensive experimentation was conducted over 130 training episodes, with the agent exposed to both step changes and varying fluctuations in ambient light. The results showed reliable convergence toward desired lighting levels with minimal training time---often under 100 milliseconds---and a moderate number of steps. Performance was evaluated using a combination of statistical summaries, histograms, and time-series plots of sensor readings, actuator outputs, and state transitions.

The system's adaptability and efficiency make it a promising solution for smart greenhouse lighting or other precision agriculture applications where maintaining optimal light conditions is critical. Importantly, the system's simplicity and autonomy allow deployment in distributed and low-power settings without the need for complex infrastructure.

Future work will expand the RL framework to simultaneously control additional environmental variables such as temperature and humidity using multi-agent architectures or hierarchical learning policies. Additionally, integrating wireless communication and cloud-based logging could further enhance scalability and remote monitoring capabilities.

\section*{Acknowledgment}

This research is based upon work supported by North Dakota State University and the U. S. Department of Agriculture, Agricultural Research Service, under agreement No. 58-6064-3-011.

\bibliographystyle{IEEEtran}
\bibliography{references}

\vspace{12pt}
\color{red}

\end{document}